\documentclass[conference,onecolumn]{IEEEtran}
\IEEEoverridecommandlockouts
\usepackage{siunitx}
\usepackage[nolist]{acronym}
\usepackage{ifthen}
\usepackage{amsmath,amssymb,amsfonts}
\usepackage{algorithmic}
\usepackage{graphicx}
\usepackage{textcomp}
\usepackage{xcolor}
\usepackage{colortbl}
\usepackage{ragged2e}
\usepackage{tabularx}
\usepackage[bookmarks=false]{hyperref}
\usepackage{comment}
\usepackage{booktabs}
\usepackage{multirow}
\usepackage{dirtytalk}
\usepackage{tikz}
\usetikzlibrary{shapes.geometric, arrows, positioning}
\usepackage{circuitikz}
\usepackage{svg}
\usepackage{pgfplots}
\usepgfplotslibrary{fillbetween}
\pgfplotsset{width=10cm,compat=1.9}
\usepackage[draft]{minted}
\usepackage[most,minted,listings]{tcolorbox}
\usepackage{caption}
\usepackage{listings}
\usepackage{wrapfig}
\usepackage{subcaption}
\usepackage{pifont}
\usepackage{soul}
\usepackage{footnote}
\usepackage{algorithm}
\usepackage{algorithmic}
\usepackage{pgfplotstable}
\usepackage{moresize}

\definecolor{ForestGreen}{RGB}{34,150,34}

\newtcolorbox{cyanbox}[2][]{
arc=4mm,
lower separated=false,
fonttitle=\bfseries,
colbacktitle=cyan!40, % different color than blue/magenta/green/yellow
coltitle=black,
enhanced,
attach boxed title to top left={xshift=0.5cm, yshift=-2mm},
colframe=gray,
colback=white,
title=#2,#1
}

\newtcolorbox{bluebox}[2][]{
arc=4mm,
lower separated=false,
fonttitle=\bfseries,
colbacktitle=blue!40,
coltitle=black,
enhanced,
attach boxed title to top left={xshift=0.5cm, yshift=-2mm},
colframe=gray,
colback=white,
title=#2,#1}

\newtcolorbox{yellowbox}[2][]{
arc=4mm,
lower separated=false,
fonttitle=\bfseries,
colbacktitle=yellow!40, % different color than blue/magenta/green/yellow
coltitle=black,
enhanced,
attach boxed title to top left={xshift=0.5cm, yshift=-2mm},
colframe=gray,
colback=white,
title=#2,#1
}

\newtcolorbox{greenbox}[2][]{
arc=4mm,
lower separated=false,
fonttitle=\bfseries,
colbacktitle=teal!50,
coltitle=black,
enhanced,
attach boxed title to top left={xshift=0.5cm, yshift=-2mm},
colframe=gray,
colback=white,
title=#2,#1}

\makeatletter
\newcommand{\linebreakand}{%
\end{@IEEEauthorhalign}
\hfill\mbox{}\par
\mbox{}\hfill\begin{@IEEEauthorhalign}
}
\makeatother
\newboolean{blindreview}
\setboolean{blindreview}{false}
\DeclareRobustCommand{\IEEEauthorrefmark}[1]{\smash{\textsuperscript{\footnotesize #1}}}
\usepackage[style=ieee, backend=biber, natbib=true, maxbibnames=1, minbibnames=1]{biblatex}
\begin{document}

%%%%%%%%%%%%%%%%%%%%%%%%%%%%%%%%%%%%%%%%%%%%%%%%%%%%%%%%%%%%%%%%%%%%%%%%%%%%%%%%%%%%%%%%
%%%%% Acronyms
%%%%%%%%%%%%%%%%%%%%%%%%%%%%%%%%%%%%%%%%%%%%%%%%%%%%%%%%%%%%%%%%%%%%%%%%%%%%%%%%%%%%%%%%
\begin{acronym}[]
    \acro{AI}[AI]{Artificial Intelligence}
    \acro{ADHD}[ADHD]{Attention Deficit Hyperactivity Disorder}
    \acro{ASIC}[ASIC]{Application Specific Integrated Circuit}
    \acro{AGI}[AGI]{Artificial General Intelligence}
    \acro{ALU}[ALU]{Arithmetic Logic Unit}
    \acro{API}{Application Programming Interface}
    \acrodefplural{CWE}[CWEs]{Common Weakness Enumerations}
    \acrodefplural{CVE}[CVEs]{Common Vulnerability Enumerations}
    \acro{CVE}[CVE]{Common Vulnerability Enumeration}
    \acro{CWE}[CWE]{Common Weakness Enumeration}
    \acro{CEX}[CEX]{Counter Example}
    \acrodefplural{CEX}[CEXs]{Counter Examples}
    \acrodefplural{CSV}[CSVs]{Comma-Separated Values}
    \acro{CoT}[CoT]{Chain-of-Thought}
    \acro{DUV}[DUV]{Design Under Verification}
    \acro{DOS}[DOS]{Denial of Service}
    \acro{DSGI}[DSGI]{Domain-Specific General Intelligence}
    \acro{EDA}[EDA]{Electronic Design Automation}
    \acro{FSM}[FSM]{Finite State Machine}
    \acrodefplural{FSM}[FSMs]{Finite State Machines}
    \acro{FIFO}[FIFO]{First-In First-Out}
    \acro{FV}[FV]{Formal Verification}
    \acro{GenAI}[GenAI]{Generative AI}
    \acro{GPT}[GPT]{Generative Pre-trained Transformer}
    \acro{HIL}[HIL]{Human-in-the-Loop}
    \acro{HDL}[HDL]{Hardware Description Language}
    \acro{IP}[IP]{Intellectual Property}
    \acrodefplural{IP}[IPs]{Intellectual Properties}
    \acro{KPI}[KPI]{Key Performance Indicator}
    \acrodefplural{KPI}[KPIs]{Key Performance Indicators}
    \acro{KG}{Knowledge Graph}
    \acrodefplural{KG}[KGs]{Knowledge Graphs}
    \acro{LLM}[LLM]{Large Language Model}
    \acrodefplural{LLM}[LLMs]{Large Language Models}
    \acro{LTLC}[LTLC]{Long Term, Long Context}
    \acro{ML}[ML]{Machine Learning}
    \acro{MMLU}[MMLU]{Massive Multitask Language Understanding}
    \acro{NLP}[NLP]{Natural Language Processing}
    \acro{PPA}[PPA]{Power, Performance and Area}
    \acro{RTL}[RTL]{Register Transfer Level}
    \acro{RAG}[RAG]{Retrieval Augmented Generation}
    \acro{RCA}[RCA]{Root Cause Analysis}
    \acro{RQ}[RQ]{Research Question}
    \acrodefplural{RQs}[RQs]{Research Questions}
    \acro{RDF}{Resource Description Framework}
    \acro{SoC}[SoC]{System-on-Chip}
    \acrodefplural{SoC}[SoCs]{System-on-Chips}
    \acro{SEU}[SEU]{Single Event Upset}
    \acrodefplural{SEU}[SEUs]{Single Event Upsets}
    \acro{SVA}[SVA]{SystemVerilog Assertion}
    \acrodefplural{SVA}[SVAs]{SystemVerilog Assertions}
    \acro{STSC}[STSC]{Short Term, Short Context}
    \acro{UVM}[UVM]{Universal Verification Methodology}
    \acro{VCD}[VCD]{Value Change Dump}
    \acro{vPlan}[vPlan]{Verification Plan}
\end{acronym}

% ~~~~~~~~~~~~~~~~~~~~~~~~~~~~~~~~~~~~~~~~~~~~~~~~~~~~~~~~~~~~~~~~~~~~~~~~
% SystemVerilog setup for listings
% ~~~~~~~~~~~~~~~~~~~~~~~~~~~~~~~~~~~~~~~~~~~~~~~~~~~~~~~~~~~~~~~~~~~~~~~~
\lstset{
    language=Verilog,           % the language of the code
    basicstyle=\footnotesize,   % the size of the fonts that are used for the code
    numbers=left,               % where to put the line-numbers; possible values are (none, left, right)
    frame=lines,                % adds a frame around the code
    captionpos=b,               % sets the caption-position to bottom
    breaklines=true,            % sets automatic line breaking
    tabsize=2,                  % sets default tabsize to 2 spaces
    xleftmargin=2.1em,
    framexleftmargin=1.7em,
    commentstyle=\color{ForestGreen},
    keywordstyle=\color{blue},
    stringstyle=\color{red},
    %linewidth=0.8\textwidth,   % define horizontal width
}

\lstdefinelanguage{Verilog}{morekeywords={accept_on,alias,always,always_comb,always_ff,always_latch,and,assert,assign,assume,automatic,before,begin,bind,bins,binsof,bit,break,buf,bufif0,bufif1,byte,case,casex,casez,cell,chandle,checker,class,clocking,cmos,config,const,constraint,context,continue,cover,covergroup,coverpoint,cross,deassign,default,defparam,design,disable,dist,do,edge,else,end,endcase,endchecker,endclass,endclocking,endconfig,endfunction,endgenerate,endgroup,endinterface,endmodule,endpackage,endprimitive,endprogram,endproperty,endspecify,endsequence,endtable,endtask,enum,event,eventually,expect,export,extends,extern,final,first_match,for,force,foreach,forever,fork,forkjoin,function,generate,genvar,global,highz0,highz1,if,iff,ifnone,ignore_bins,illegal_bins,implements,implies,import,incdir,include,initial,inout,input,inside,instance,int,integer,interconnect,interface,intersect,join,join_any,join_none,large,let,liblist,library,local,localparam,logic,longint,macromodule,matches,medium,modport,module,nand,negedge,nettype,new,nexttime,nmos,nor,noshowcancelled,not,notif0,notif1,null,or,output,package,packed,parameter,pmos,posedge,primitive,priority,program,property,protected,pull0,pull1,pulldown,pullup,pulsestyle_ondetect,pulsestyle_onevent,pure,rand,randc,randcase,randsequence,rcmos,real,realtime,ref,reg,reject_on,release,repeat,restrict,return,rnmos,rpmos,rtran,rtranif0,rtranif1,s_always,s_eventually,s_nexttime,s_until,s_until_with,scalared,sequence,shortint,shortreal,showcancelled,signed,small,soft,solve,specify,specparam,static,string,strong,strong0,strong1,struct,super,supply0,supply1,sync_accept_on,sync_reject_on,table,tagged,task,this,throughout,time,timeprecision,timeunit,tran,tranif0,tranif1,tri,tri0,tri1,triand,trior,trireg,type,typedef,union,unique,unique0,unsigned,until,until_with,untyped,use,uwire,var,vectored,virtual,void,wait,wait_order,wand,weak,weak0,weak1,while,wildcard,wire,with,within,wor,xnor,xor,`uvm_create, `uvm_rand_send_with},morecomment=[l]{//}}

\title{Saarthi for AGI: Towards Domain-Specific General Intelligence for Formal Verification\\

}

%%%%%%%%%%%%%%%%%%%%%%%%%%%%%%%%%%%%%%%%%%%%%%%%%%%%%%%%%%%%%%%%%%%%%%%%%%%%%%%%%%%%%%%%
%%%%% Authors
%%%%%%%%%%%%%%%%%%%%%%%%%%%%%%%%%%%%%%%%%%%%%%%%%%%%%%%%%%%%%%%%%%%%%%%%%%%%%%%%%%%%%%%%
\ifthenelse{\boolean{blindreview}}{}{
	\author{\IEEEauthorblockN{
	        Aman Kumar\IEEEauthorrefmark{1},
            Deepak Narayan Gadde\IEEEauthorrefmark{2},
            Luu Danh Minh\IEEEauthorrefmark{3},\\
            Vaisakh Naduvodi Viswambharan\IEEEauthorrefmark{2},
			Keerthan Kopparam Radhakrishna\IEEEauthorrefmark{2},
			Sivaram Pothireddypalli\IEEEauthorrefmark{1}}
		\IEEEauthorblockA{
			\IEEEauthorrefmark{1}Infineon Technologies India Private Limited, India \\
            \IEEEauthorrefmark{2}Infineon Technologies Dresden AG \& Co. KG, Germany \\
            \IEEEauthorrefmark{3}Infineon Technologies Vietnam Company Ltd., Vietnam}
        %\textit{E-mail: Firstname.Lastname@infineon.com}
	}
}
%%%%%%%%%%%%%%%%%%%%%%%%%%%%%%%%%%%%%%%%%%%%%%%%%%%%%%%%%%%%%%%%%%%%%%%%%%%%%%%%%%%%%%%%

\maketitle

\begin{abstract}
\textbf{Saarthi~\cite{saarthi} is an agentic AI framework that uses multi-agent collaboration to perform end-to-end formal verification. Even though the framework provides a complete flow from specification to coverage closure, with around \SI{40}{\percent} efficacy, there are several challenges that need to be addressed to make it more robust and reliable. \ac{AGI} is still a distant goal, and current \ac{LLM}-based agents are prone to hallucinations and making mistakes, especially when dealing with complex tasks such as formal verification. However, with the right enhancements and improvements, we believe that Saarthi can be a significant step towards achieving domain-specific general intelligence for formal verification. Especially for problems that require \ac{STSC} capabilities, such as formal verification, Saarthi can be a powerful tool to assist verification engineers in their work. In this paper, we present two key enhancements to the Saarthi framework: (1) a structured rulebook and specification grammar to improve the accuracy and controllability of \ac{SVA} generation, and (2) integration of advanced \ac{RAG} techniques, such as GraphRAG~\cite{graphrag}, to provide agents with access to technical knowledge and best practices for iterative refinement and improvement of outputs. We also benchmark these enhancements for the overall Saarthi framework using challenging test cases from NVIDIA's CVDP benchmark~\cite{nvidia_cvdp} targeting formal verification. Our benchmark results stand out with a \SI{70}{\percent} improvement in the accuracy of generated assertions, and a \SI{50}{\percent} reduction in the number of iterations required to achieve coverage closure.}
\end{abstract}


\section{Introduction}  \label{intro}

The increasing complexity, configurability, and safety-criticality of modern semiconductor designs have intensified demands on \ac{AI} workflows. Industry-standard methodologies rely on expert engineers to interpret specifications, distill them into verification plans, author and refine \acp{SVA}, and iteratively close proof, coverage, and vacuity gaps. While \ac{FV} offers exhaustive guarantees, it remains labor-intensive and difficult to scale. Advances in \acp{LLM} and agentic \ac{AI} suggest potential for automating substantial portions of this pipeline \cite{saarthi,genai_sst}. However, limitations such as syntactic instability, semantic misinterpretation, and shallow reasoning persist \cite{reformai,10458102}.

Emerging work on multi-agent orchestration \cite{saarthi,gadde2025heyaigeneratehardware} demonstrates that decomposing \ac{FV} into specialized roles improves robustness. Yet, gaps remain in assertion synthesis controllability, grounding agent reasoning in authoritative corpora, and adaptive feedback mechanisms. Addressing these gaps is essential for progress toward \ac{DSGI} in \ac{FV}—a tractable milestone on the trajectory toward \ac{AGI} \cite{agi_spark,situational_awareness}.

\ac{RAG} \cite{lewis2021retrievalaugmentedgenerationknowledgeintensivenlp} mitigates hallucinations and increases factual precision by conditioning generation on retrieved evidence. In \ac{EDA}, authoritative sources include IEEE SystemVerilog standards \cite{10458102}, ISA specifications \cite{riscv_ratified_specifications}, and prior assertion libraries. GraphRAG \cite{graphrag} extends \ac{RAG} by retrieving structured subgraphs, enabling multi-hop consistency checks and traceability.

This paper advances \ac{AI}-driven \ac{FV} through: (1) a structured rulebook and specification grammar for \ac{SVA} generation; (2) integration of advanced \ac{RAG} techniques, such as GraphRAG \cite{graphrag}; (3) automated coverage hole-filling; (4) \ac{HIL} data collection pipelines; and (5) benchmarking on diverse designs. Enhancements yield up to a \SI{70}{\percent} increase in assertion accuracy and a \SI{50}{\percent} reduction in iterations to coverage closure. By coupling controllable generation with structured knowledge grounding and experiential learning, Saarthi narrows the reliability gap in domain-specialized cognitive workflows, advancing \ac{DSGI}.

\section{Background} \label{bgnd}

This paper represents a significant step toward achieving \ac{AGI} in the domain of formal verification. By leveraging the capabilities of \acp{LLM} and integrating advanced techniques such as multi-agent collaboration, structured rulebooks, and retrieval-augmented generation, we aim to address the challenges of automating complex verification workflows. Unlike prior work that highlighted the limitations of \ac{LLM}-generated outputs \cite{reformai}, our approach focuses on iterative refinement, grounding in authoritative knowledge, and systematic feedback loops to enhance reliability and accuracy. Use cases, such as formal verification, that require \ac{STSC} capabilities for \acp{LLM} are well-suited to the path toward achieving \ac{AGI} \cite{ai_hype}. These efforts align with the broader vision of \ac{DSGI}, where domain-specific intelligence can tackle intricate tasks like formal verification, paving the way for scalable and robust \ac{AI}-driven engineering solutions.

\subsection{Retrieval-Augmented Generation (RAG)}

\ac{RAG} integrates information retrieval with neural text generation, enabling models to condition outputs on external sources for accurate, context-aware results \cite{lewis2021retrievalaugmentedgenerationknowledgeintensivenlp, izacard2021leveragingpassageretrievalgenerative}. In hardware verification, engineers consult standards like IEEE SystemVerilog \cite{10458102}, \ac{UVM} \cite{9195920}, and ISA manuals \cite{riscv_ratified_specifications}. Grounding responses in cited passages, \ac{RAG} improves factuality and reduces hallucinations, supporting traceable workflows \cite{ISO26262_1_2018, RTCA_DO254_2000}.

\begin{figure}[h!]
\centering
  \includegraphics [width=0.5\textwidth] {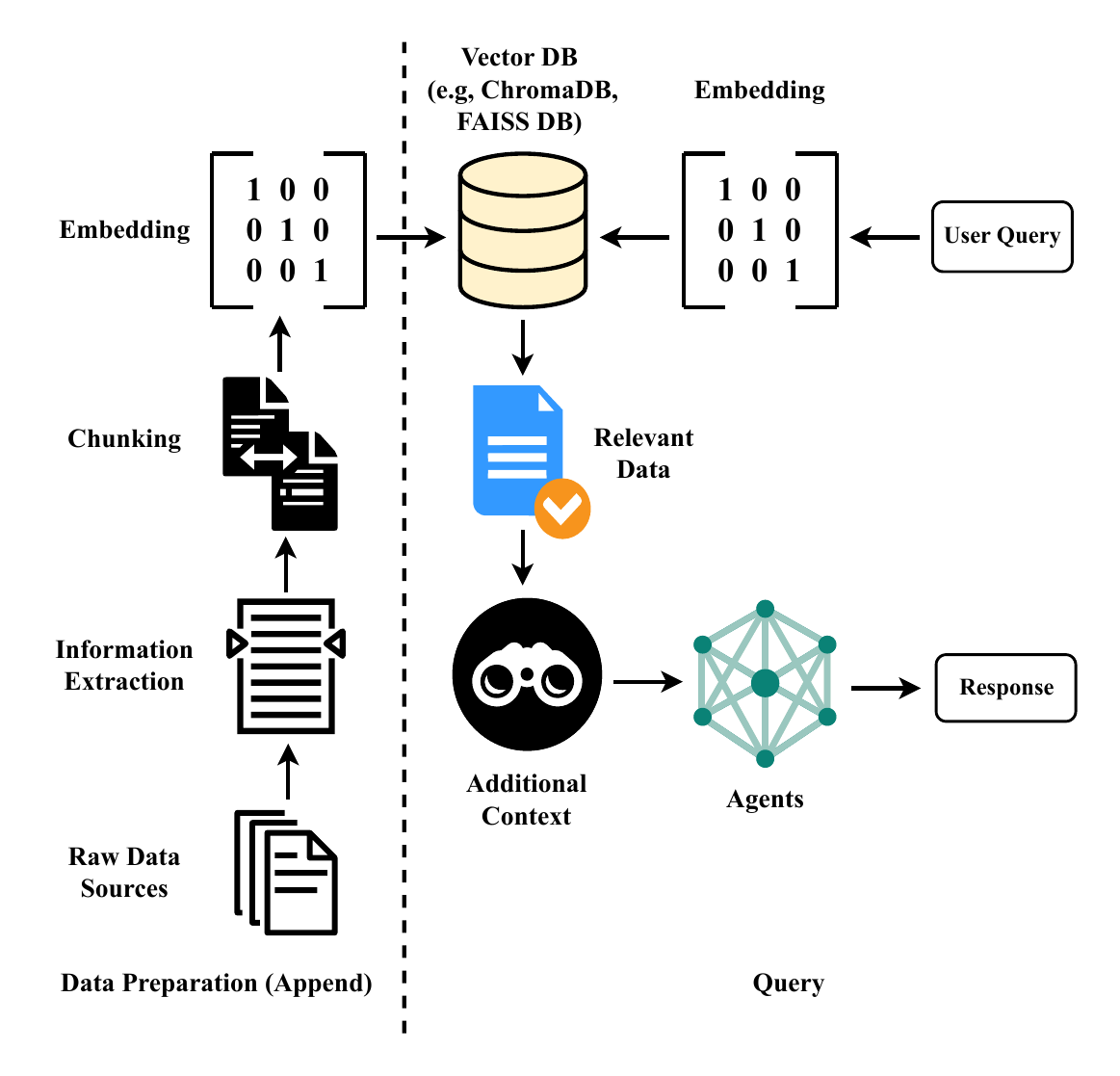}
\caption{Overview of the basic \ac{RAG} pipeline \cite{ImplementingLangChain}}
\label{fig:ragfaiss}
\end{figure}

A \ac{RAG} pipeline combines offline indexing of domain corpora with hybrid retrieval and grounded generation. The corpus includes standards, repositories, and logs. Documents are chunked, embedded, and stored in an index (e.g., Facebook AI Similarity Search (FAISS)) \cite{johnson2017billionscalesimilaritysearchgpus} as shown in Figure~\ref{fig:ragfaiss}. At query time, a hybrid retriever pairs dense retrieval with lexical matching, reranking top candidates for precision \cite{karpukhin2020densepassageretrievalopendomain}. The generator produces grounded outputs (e.g., \acp{SVA} with citations) \cite{lewis2021retrievalaugmentedgenerationknowledgeintensivenlp}. Enhancements like query reformulation \cite{gao2022precisezeroshotdenseretrieval} and iterative reasoning \cite{yao2023reactsynergizingreasoningacting} further improve results.

\subsection{Knowledge Graph}

A \ac{KG} represents entities as nodes and their relationships as edges, structured by a schema or ontology \cite{10.1145/3447772,defkg}. \acp{KG} can be implemented as \ac{RDF} triples or property graphs (e.g., Neo4j) and accessed via query languages like SPARQL or Cypher \cite{rdfconcepts,w3SPARQLQuery,oreillyGraphDatabases}. They enable precise querying, inference, and integration with vector-based similarity measures \cite{10.1145/3447772,10.3233/SW-160218}.

Standard \ac{KG} workflows unify diverse data sources (e.g., documents, logs) into a graph through entity and relationship extraction, schema alignment, and enhancement \cite{10.1145/3447772,10.3233/SW-160218}. \acp{KG} support multi-hop queries, provenance tracing, and analytics, facilitating traceability in hardware verification \cite{10458102,9195920,eurocaeED80Design,ISO26262_1_2018}.

In verification, \ac{KG}-based traceability links requirements to \ac{RTL} modules, assertions, tests, and coverage, enabling queries like identifying unvalidated requirements or correlating failures with design hierarchies. GraphRAG enhances retrieval by leveraging \ac{KG} topology for multi-hop reasoning, improving factual consistency and recall on long-horizon dependencies \cite{lewis2021retrievalaugmentedgenerationknowledgeintensivenlp,githubGitHubMicrosoftgraphrag}.

\subsection{GraphRAG}
\begin{figure}[h!]
\centering
  \includegraphics [width=0.6\textwidth] {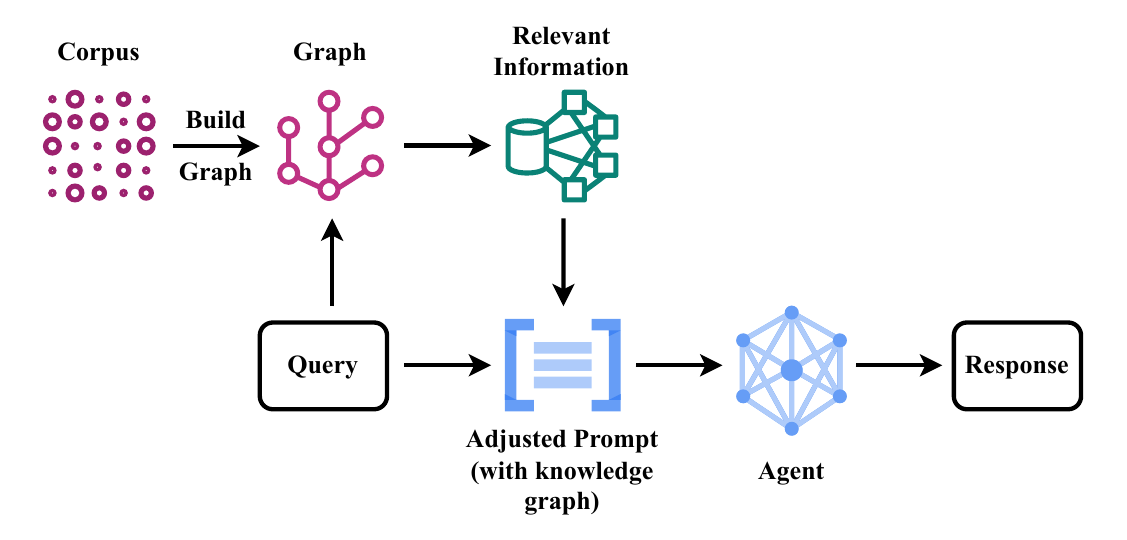}
\caption{Knowledge graph construction and retrieval in GraphRAG pipeline \cite{rabilooGraphRAG}}
\label{fig:graphrag}
\end{figure}

GraphRAG is a \ac{RAG} approach that uses a knowledge graph as its evidence source, as illustrated in Figure~\ref{fig:graphrag}. The pipeline begins by processing a corpus of unstructured documents to construct a knowledge graph, where nodes represent formal verification artifacts (requirements, signals, properties, proofs, counterexamples) and edges capture relations (e.g., \emph{implements}, \emph{constrains}, \emph{proven\_by}, \emph{violated\_by}). When a query arrives, the system retrieves a relevant subgraph containing interconnected information rather than isolated text fragments. This structured evidence is then integrated into the prompt context for the \ac{LLM}. The model leverages both the query and the graph-derived knowledge to generate responses grounded in relational reasoning, enabling multi-hop inference and producing more accurate, contextually coherent answers than traditional chunk-based retrieval.

\textbf{Example:} Suppose you ask, \say{How should I check that \lstinline{AXI} \lstinline{WLAST} matches the \lstinline{AWLEN} burst length?}. The graph links \lstinline{AWLEN} to the rule burst\_length = \lstinline{AWLEN} + 1, \lstinline{WLAST} to ``asserted on the last data beat,'' and the handshake relation that each \lstinline{WVALID && WREADY} advances the beat count. Graph \ac{RAG} retrieves this subgraph and produces a simple check:

\begin{lstlisting}
After AWVALID && AWREADY handshake
Count data-beat handshakes
Check: WLAST = 1 on (AWLEN + 1)-th beat
       WLAST = 0 before that
Assume: AWLEN stable during burst
\end{lstlisting}

The answer cites the specific nodes (\lstinline{AWLEN}, \lstinline{WLAST}, handshake rule), preserving provenance for auditability.

\section{Saarthi: Agentic \ac{AI}-Based Formal Verification} \label{setup}

\begin{comment}
\begin{figure}[h!]
\centering
  \includegraphics [width=0.75\textwidth] {Images/formal_flow.pdf}
\caption{Saarthi: Agentic \ac{AI} based formal verification using multi-agent collaboration}
\label{formal_flow}
\end{figure}
\end{comment} 
The framework presented in \cite{gadde2025heyaigeneratehardware} \cite{saarthi} laid the groundwork for \ac{AI}-driven formal verification, emphasizing agent coordination, \ac{EDA} tool integration, and the potential for explainable \ac{AI} in hardware design workflows. Extending this foundation, the present work advances the methodology through three primary directions. First, a structured rulebook and specification grammar are introduced to enhance the accuracy and controllability of \ac{SVA} generation. Second, advanced \ac{RAG} approaches, such as GraphRAG \cite{graphrag}, are incorporated to improve the agents’ access to technical knowledge and iterative reasoning capabilities. Third, an automated coverage hole–filling mechanism is proposed to identify and generate targeted assertions that address unverified design regions, thereby accelerating coverage closure. Together, these enhancements significantly strengthen the robustness and practical applicability of the Saarthi framework, as validated through benchmarks on NVIDIA’s Comprehensive Verilog Design Problems (CVDP) dataset \cite{nvidia_cvdp}.

To realize our contributions and conduct the experiments, we implemented the flow shown in Figure~\ref{fig:formal_flow}. Upon task assignment, \ac{AI} agents assume primary control of the verification workflow. Saarthi facilitates formal verification through a multi-agent, agentic \ac{AI} approach that coordinates specialized agents. The framework incorporates design patterns for agentic reasoning and safeguards to mitigate context limitations, hallucinations, and repetitive loops. Saarthi is built on Microsoft AutoGen, leveraging its multi-agent orchestration capabilities to support formal verification. Its architecture provides a configurable orchestration layer that can be tailored to diverse verification requirements while preserving process consistency and reliability.

\subsection{Agent Orchestration} 

The orchestrator supports both sequential and hierarchical execution; for this formal verification process, agents are arranged sequentially. After orchestration, the selected framework’s main module initializes the verification run and invokes the agents in order.

During this process, the agents generate key artifacts such as \ac{vPlan} and properties, logging their interactions and collaborations as they proceed. Generated properties are evaluated by critic agents, who provide feedback to improve the accuracy and correctness of \acp{SVA}. This iterative loop continues until a convergence threshold is reached. If the agents cannot finalize the \acp{SVA} within a predefined iteration limit, human intervention (i.e., \ac{HIL}) is triggered for further assessment. Once finalized, the \acp{SVA} undergo formal verification, and any \acfp{CEX} are identified, and resolved by the agents.

%\begin{comment}
%%%DEEPAK is editing%%% 

\subsection{Multi-Agent Workflow for \ac{AI}-Driven Formal Verification}

\begin{figure}[h!]
\centering
  \includegraphics [width=0.6\textwidth] {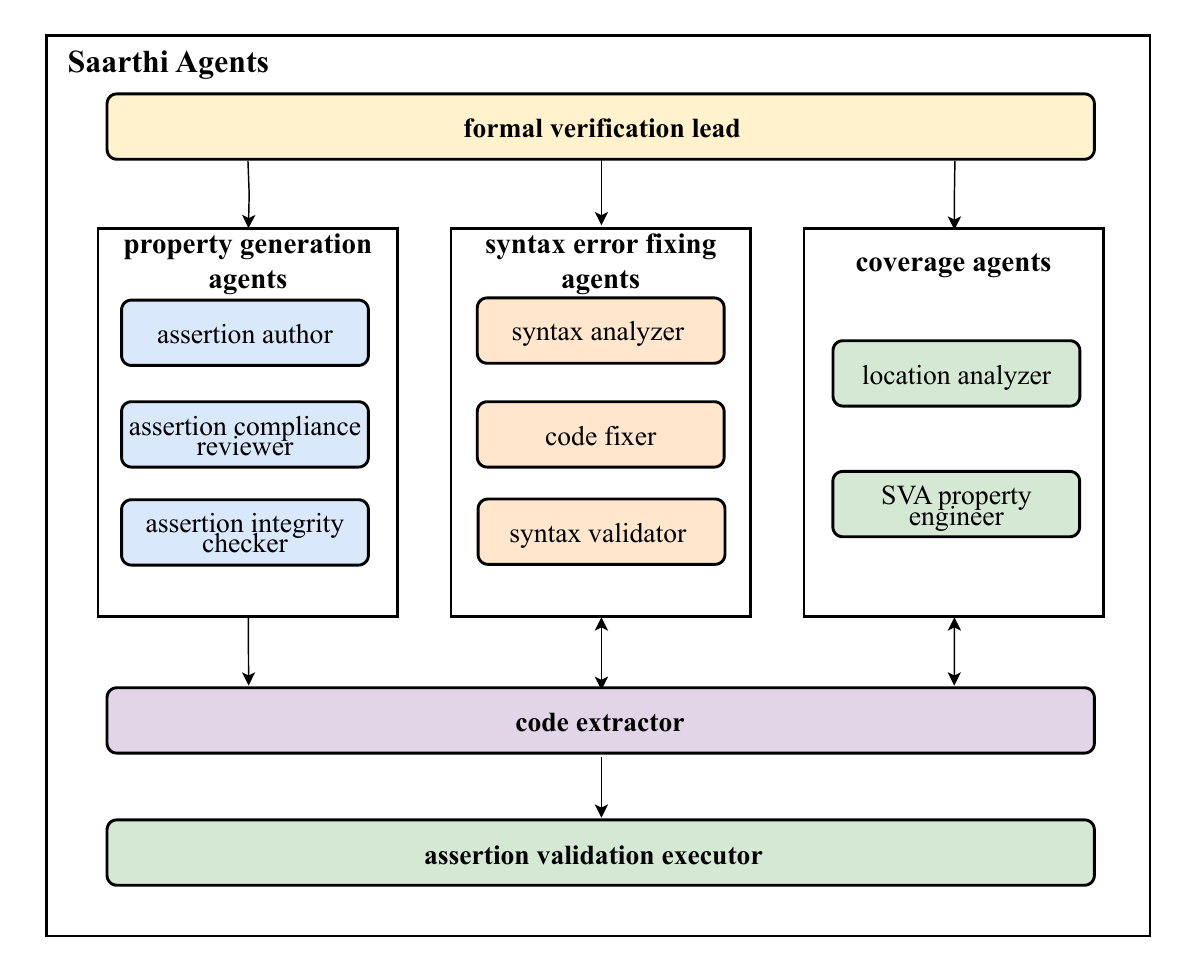}
\caption{Saarthi: Agentic \ac{AI} based formal verification using multi-agent collaboration}
\label{fig:formal_flow}
\end{figure}

To implement our contributions and execute our experiments, we developed the multi-agent workflow illustrated in Figure~\ref{fig:formal_flow}. The framework orchestrates the formal verification process through a coordinator/lead that manages the end-to-end formal verification plan for the \ac{DUV}.

\textit{Property-generation agents} systematically translate functional requirements and protocol rules into candidate \acp{SVA}, annotating each assertion with metadata including the target module, interface, and signal bindings. \textit{Syntax-error-fixing agents} sanitize these candidates through a three-stage pipeline: a syntax analyzer detects parser and lint violations, a code fixer applies canonical rewrites and enforces naming conventions, and a syntax validator verifies compliance against tool-chain compilers. \textit{Coverage agents} perform gap analysis and strategic placement: a location analyzer identifies optimal insertion points within the \ac{RTL} hierarchy, while an \ac{SVA} property engineer synthesizes additional assertions to address uncovered behaviors. This decomposition enables concurrent progress across property authoring, correction, and placement phases, with clear ownership boundaries and coordinator/lead oversight.

A code extractor/manager mediates interactions among agents, tools, and human reviewers. Its primary function is to materialize agent-generated code and update the design repository and/or verification environment with the extracted artifacts. The extractor returns integration-ready deliverables and performs binding into the verification environment or specific modules.

Coverage feedback and tool diagnostics drive the next iteration. Coverage agents analyze reports to identify gaps and missing checks, then issue new property requests with target modules and signals. Compile errors and vacuity warnings are routed back to the syntax-fixing loop for correction and revalidation.

Each cycle repairs the code, normalizes naming, updates the code index/graph, and re-runs proofs. This reduces authoring time, minimizes pre-run syntax churn, accelerates coverage closure, maintains correct assertion placement, and preserves provenance for auditability and reproducibility.

\subsection{Dataset Collection during Human-in-the-Loop (\ac{HIL}) Refinement}

\begin{figure}[h!]
\centering
  \includegraphics [width=0.4\textwidth] {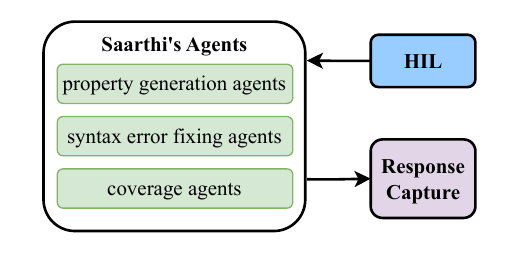}
\caption{Data Collection Flow during \ac{HIL}}
\label{fig:hil_response}
\end{figure}

During the \ac{HIL} phase, we systematically collect the validated response set produced following human interventions as shown in Figure~\ref{fig:hil_response}. Each corrected or approved agent output is logged into a structured dataset that records the refined response, its originating prompt and context, observed error signatures (e.g., parse failures, vacuity, misbinding), and the applied resolution pattern. This collection ensures that every human‑validated example contributes to a growing corpus of high‑quality artifacts reflecting real failure recoveries. The resulting dataset underpins iterative improvement enabling prompt adjustments, template extensions for previously uncovered properties, and targeted repair of syntactic and semantic defects identified during the \ac{HIL} cycle. By continuously enriching this corpus, the framework builds a self‑improving feedback base that increases agents' reliability and reduces the need for manual intervention over time.

Furthermore, the framework supports standalone invocation of individual agents without executing the full workflow to address specific use cases or targeted applications.

\subsection{Rulebooks}

Inspired by the BugGen fault injection methodology~\cite{jasper2025buggen}, we propose a structured rulebook to standardize specification grammar in assertion generation workflows. This rulebook bridges ambiguous natural language descriptions and machine-readable formats, enhancing automation and interpretability.

Unlike prior approaches relying on natural language instructions, our methodology encodes specifications as concise keywords, creating a deterministic pipeline for assertion generation. This uniform representation improves predictability and facilitates debugging by human engineers.

\begin{comment}
Inspired by the BugGen fault injection methodology~\cite{jasper2025buggen}, we propose a structured rulebook designed to standardize the specification grammar used in assertion generation workflows. This rulebook serves as a bridge between ambiguous natural language descriptions and a discrete, machine-readable format, thereby improving both automation and interpretability in the design verification process.

In contrast to our previous approach, where a manager agent issued instructions in natural language, the revised methodology encodes specifications as a concise array of keywords. These keywords enable more direct context cue between the specification and the design \ac{RTL}, enabling a more deterministic and streamlined pipeline for assertion generation. This uniform representation not only enhances the predictability but also facilitates manual debugging by human engineers, who benefit from the clarity and consistency of the format.
\end{comment}

\begin{comment}
\begin{figure}[h!]
\centering
  \includegraphics [width=0.5\textwidth] {Images/rulebook_cache.pdf}
\caption{A snippet of the cached mistakes that all agents can learn from for future iterations}
\label{fig:rulebook_cache}
\end{figure}
\end{comment}

\begin{figure}[h!]
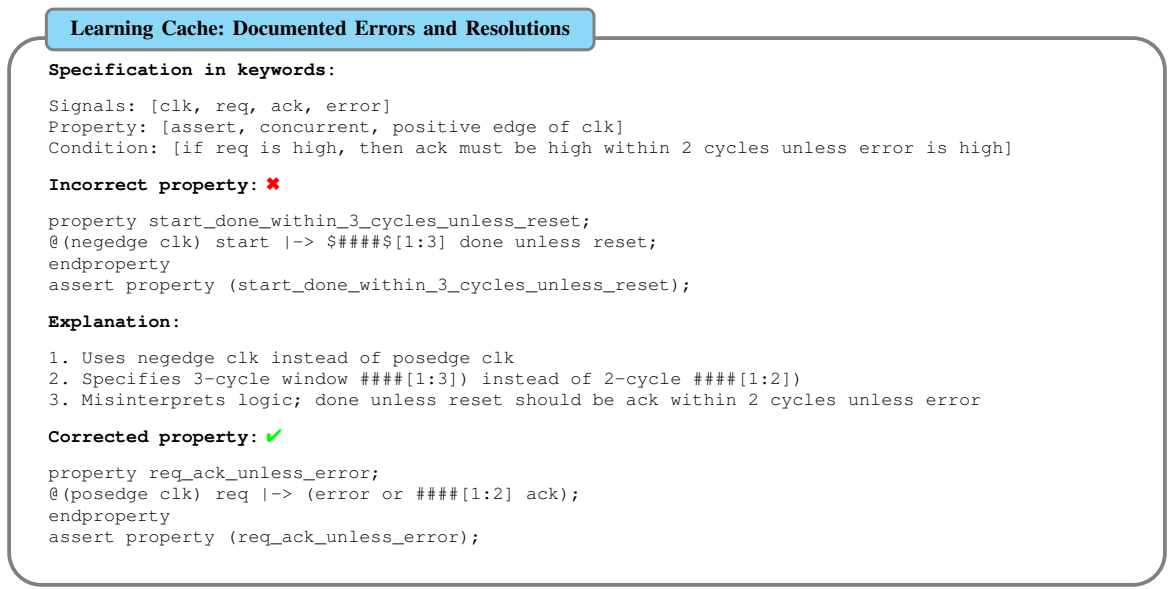

\centering
\begin{cyanbox}[width=0.85\linewidth]{\footnotesize Learning Cache: Documented Errors and Resolutions}
{\scriptsize
{\ttfamily\bfseries\detokenize{Specification in keywords:}}\par
\vspace{0.2cm}
{\ttfamily\detokenize{Signals: [clk, req, ack, error]}}\par
{\ttfamily\detokenize{Property: [assert, concurrent, positive edge of clk]}}\par
{\ttfamily\detokenize{Condition: [if req is high, then ack must be high within 2 cycles unless error is high]}}\par
\medskip
{\ttfamily\bfseries\detokenize{Incorrect property:}} \textcolor{red}{\ding{54}}\par
\vspace{0.2cm}
{\ttfamily\detokenize{property start_done_within_3_cycles_unless_reset;}}\par
{\ttfamily\detokenize{@(negedge clk) start |-> $##$[1:3] done unless reset;}}\par
{\ttfamily\detokenize{endproperty}}\par
{\ttfamily\detokenize{assert property (start_done_within_3_cycles_unless_reset);}}\par
\medskip
{\ttfamily\bfseries\detokenize{Explanation:}}\par
\vspace{0.2cm}
{\ttfamily\detokenize{1. Uses negedge clk instead of posedge clk}}\par
{\ttfamily\detokenize{2. Specifies 3-cycle window ##[1:3]) instead of 2-cycle ##[1:2])}}\par
{\ttfamily\detokenize{3. Misinterprets logic; done unless reset should be ack within 2 cycles unless error}}\par
\medskip
{\ttfamily\bfseries\detokenize{Corrected property:}} \textcolor{green}{\ding{52}}\par
\vspace{0.2cm}
{\ttfamily\detokenize{property req_ack_unless_error;}}\par
{\ttfamily\detokenize{@(posedge clk) req |-> (error or ##[1:2] ack);}}\par
{\ttfamily\detokenize{endproperty}}\par
{\ttfamily\detokenize{assert property (req_ack_unless_error);}}\par
\medskip
}
\end{cyanbox}
\caption{A snippet of the cached mistakes that all agents can learn from for future iterations}
\label{fig:rulebook_cache}
\end{figure}

The rulebook includes best practices and common pitfalls, documented with corrections and explanations (Figure~\ref{fig:rulebook_cache}). These insights assist agents and engineers in avoiding recurring errors. The rulebook is transferable across hardware designs, allowing design-specific details to be appended.

\begin{figure}[h!]
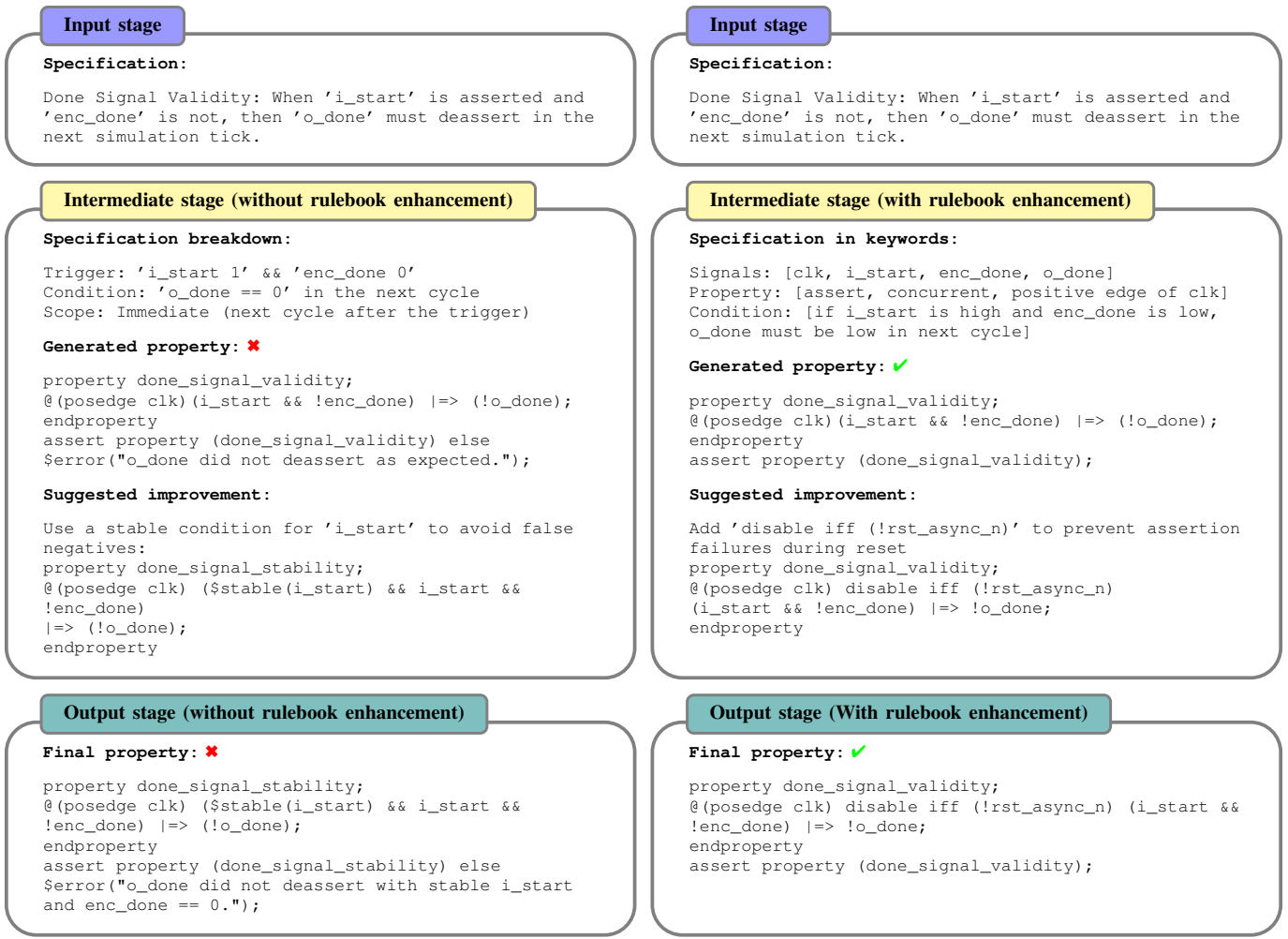

\centering

% Row 1: blue boxes
\begin{tcbraster}[raster columns=2,raster equal height]
\begin{bluebox}[]{\footnotesize Input stage}
{\scriptsize
{\ttfamily\bfseries\detokenize{Specification:}}\par
\vspace{0.2cm}
{\ttfamily\detokenize{Done Signal Validity: When 'i_start' is asserted and 'enc_done' is not, then 'o_done' must deassert in the next simulation tick.}}\par
}
\end{bluebox}
\begin{bluebox}[]{\footnotesize Input stage}
{\scriptsize
{\ttfamily\bfseries\detokenize{Specification:}}\par
\vspace{0.2cm}
{\ttfamily\detokenize{Done Signal Validity: When 'i_start' is asserted and 'enc_done' is not, then 'o_done' must deassert in the next simulation tick.}}\par
}
\end{bluebox}
\end{tcbraster}

% Row 2: magenta boxes
\begin{tcbraster}[raster columns=2,raster equal height]
\begin{yellowbox}[]{\footnotesize Intermediate stage (without rulebook enhancement)}
{\scriptsize
{\ttfamily\bfseries\detokenize{Specification breakdown:}}\par
\vspace{0.2cm}
{\ttfamily\detokenize{Trigger: 'i_start 1' && 'enc_done 0'}}\par
{\ttfamily\detokenize{Condition: 'o_done == 0' in the next cycle}}\par
{\ttfamily\detokenize{Scope: Immediate (next cycle after the trigger)}}\par
\medskip
{\ttfamily\bfseries\detokenize{Generated property:}} {\textcolor{red}{\ding{54}}\par
\vspace{0.2cm}
{\ttfamily\detokenize{property done_signal_validity;}}\par
{\ttfamily\detokenize{@(posedge clk)(i_start && !enc_done) |=> (!o_done);}}\par
{\ttfamily\detokenize{endproperty}}\par
{\ttfamily\detokenize{assert property (done_signal_validity) else $error("o_done did not deassert as expected.");}}\par
\medskip
{\ttfamily\bfseries\detokenize{Suggested improvement:}}\par
\vspace{0.2cm}
{\ttfamily\detokenize{Use a stable condition for 'i_start' to avoid false negatives:}}\par
{\ttfamily\detokenize{property done_signal_stability;}}\par
{\ttfamily\detokenize{@(posedge clk) ($stable(i_start) && i_start && !enc_done)}}\par
{\ttfamily\detokenize{|=> (!o_done);}}\par
{\ttfamily\detokenize{endproperty}}\par
}}
\end{yellowbox}
\begin{yellowbox}[]{\footnotesize Intermediate stage (with rulebook enhancement)}
{\scriptsize
{\ttfamily\bfseries\detokenize{Specification in keywords:}}\par
\vspace{0.2cm}
{\ttfamily\detokenize{Signals: [clk, i_start, enc_done, o_done]}}\par
{\ttfamily\detokenize{Property: [assert, concurrent, positive edge of clk]}}\par
{\ttfamily\detokenize{Condition: [if i_start is high and enc_done is low, o_done must be low in next cycle]}}\par
\medskip
{\ttfamily\bfseries\detokenize{Generated property:}} {\textcolor{green}{\ding{52}}\par
\vspace{0.2cm}
{\ttfamily\detokenize{property done_signal_validity;}}\par
{\ttfamily\detokenize{@(posedge clk)(i_start && !enc_done) |=> (!o_done);}}\par
{\ttfamily\detokenize{endproperty}}\par
{\ttfamily\detokenize{assert property (done_signal_validity);}}\par
\medskip
{\ttfamily\bfseries\detokenize{Suggested improvement:}}\par
\vspace{0.2cm}
{\ttfamily\detokenize{Add 'disable iff (!rst_async_n)' to prevent assertion failures during reset}}\par
{\ttfamily\detokenize{property done_signal_validity;}}\par
{\ttfamily\detokenize{@(posedge clk) disable iff (!rst_async_n)}}\par
{\ttfamily\detokenize{(i_start && !enc_done) |=> !o_done;}}\par
{\ttfamily\detokenize{endproperty}}\par
}}
\end{yellowbox}
\end{tcbraster}

% Row 3: green boxes
\begin{tcbraster}[raster columns=2,raster equal height]
\begin{greenbox}[]{\footnotesize Output stage (without rulebook enhancement)}
{\scriptsize
{\ttfamily\bfseries\detokenize{Final property:}} {\textcolor{red}{\ding{54}}\par
\vspace{0.2cm}
{\ttfamily\detokenize{property done_signal_stability;}}\par
{\ttfamily\detokenize{@(posedge clk) ($stable(i_start) && i_start && !enc_done) |=> (!o_done);}}\par
{\ttfamily\detokenize{endproperty}}\par
{\ttfamily\detokenize{assert property (done_signal_stability) else $error("o_done did not deassert with stable i_start and enc_done == 0.");}}\par
} }
\end{greenbox}
\begin{greenbox}[]{\footnotesize Output stage (With rulebook enhancement)}
{\scriptsize
{\ttfamily\bfseries\detokenize{Final property:}} {\textcolor{green}{\ding{52}}\par
\vspace{0.2cm}
{\ttfamily\detokenize{property done_signal_validity;}}\par
{\ttfamily\detokenize{@(posedge clk) disable iff (!rst_async_n) (i_start && !enc_done) |=> !o_done;}}\par
{\ttfamily\detokenize{endproperty}}\par
{\ttfamily\detokenize{assert property (done_signal_validity);}}\par
} }
\end{greenbox}
\end{tcbraster}

\caption{A comparison of the iterative \ac{SVA} generation process before and after adding the rulebook}
\label{fig:rulebook_enhancement}
\end{figure}

Figure~\ref{fig:rulebook_enhancement} illustrates the impact of structured grammar. Without it, manager agent notes vary, causing instability. Structured grammar ensures consistent assertion generation, enabling engineers to refine specifications rather than adjust prompts. Cached mistakes, such as handling reset conditions in assertions, are reused to enhance reliability and reduce debugging efforts.

Ultimately, the rulebook improves accuracy, controllability, and collaboration in \ac{SVA} generation, creating a robust and scalable solution.

\begin{comment}
Ultimately, our goal is to improve the accuracy, controllability, and collaborative potential of the \ac{SVA} generation process. By establishing a shared framework that supports both automated agents and human contributions, a more robust and scalable solution is within reach.
\end{comment}

%\subsection{RAG} %%% GraphRAG Deepak 

\subsection{\ac{RCA}} %%% Keerthan
\begin{figure}[ht!]
\centering
  \includegraphics [width=0.7\textwidth] {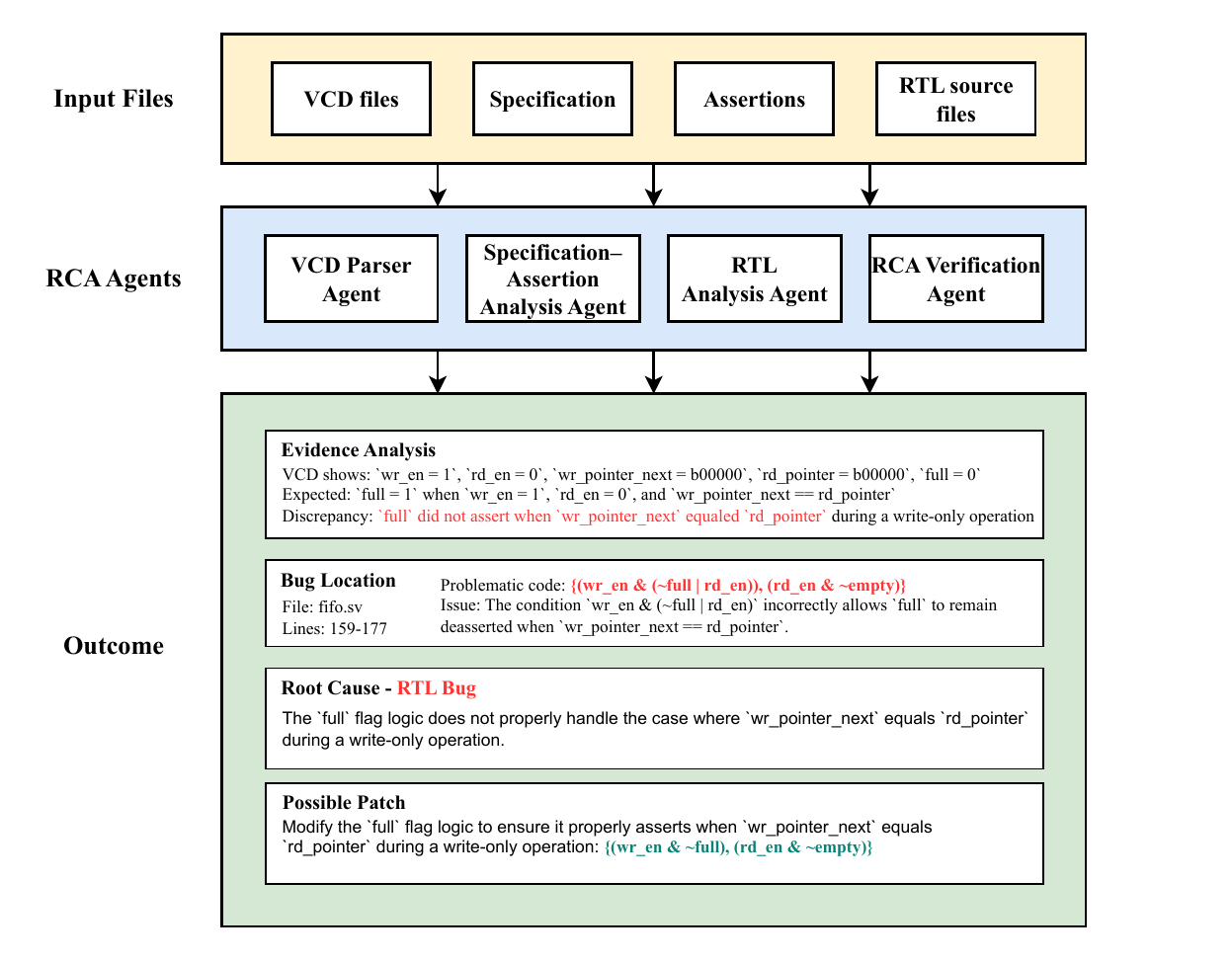}
\caption{Overview of the comprehensive multi-agent \ac{RCA} setup}
\label{fig:rca}
\end{figure}

Diagnosing and resolving \acp{CEX} is a critical bottleneck in formal verification. To address this, we deploy multi-agent \ac{RCA} agents that analyze failures and propose corrections collaboratively, as shown in Figure~\ref{fig:rca}.

Upon property violation, the formal engine generates \ac{VCD} traces capturing signal activity. These, along with the specification, \ac{RTL} sources, and failing \acp{SVA}, serve as inputs to four specialized agents: the \ac{VCD} Parser extracts signal values and timestamps; the Specification–Assertion Analyst verifies property correctness; the \ac{RTL} Analysis Agent diagnoses signal dependencies and logic conditions; and the Verification Agent ensures consistency across analyses.

The agents iteratively refine diagnoses, generating structured reports with evidence analysis, bug locations, root cause classifications, and proposed patches. If unresolved after three iterations, the framework transitions to \ac{HIL} mode, where human engineers review and validate the findings. This hybrid model balances automation with expert oversight for complex cases.

\section{Benchmarking and Results} \label{eval}

To evaluate performance and benchmark capabilities, we used the new Saarthi to verify \ac{RTL} designs of varying complexity. Alongside our in-house designs, ECC and Automotive \ac{IP}, we included three publicly available designs: Memory Scheduler, AXI4Lite, and CIC Decimator, sourced from the NVIDIA CVDP agentic \ac{AI} benchmark for assertion generation~\cite{nvidiacvdp}. For additional variety, we also incorporated a floating-point multiplier design from~\cite{tsarnadelis2023hw2project}. The agents utilized four models: GPT-4.1 and GPT-5 from OpenAI~\cite{openai2024gpt4} and LLama 3.3 from Meta~\cite{meta2024llama3_3}.

This paper evaluates performance across three comprehensive benchmarks: \acfp{KPI}, \ac{HIL} vs. No \ac{HIL}, and fully automated coverage improvement. We adopt the same \acp{KPI} from our previous work~\cite{saarthi}, with two additions: \textit{first generation success} and \textit{number of fix attempts after failure}. First generation success measures whether the generated assertions compile and run correctly on the first attempt, indicating the effectiveness of our assertion rulebooks in minimizing syntax errors. If the initial attempt fails, the number of fix attempts tracks how many iterations the syntax-fixing agent requires to produce a valid result, with a maximum of five attempts. The second benchmark compares assertion generation with and without human feedback. In the \ac{HIL} setting, after the coverage improvement agent runs and initial results are recorded (see the left side of Table~\ref{hil_coverage_results}), a human engineer provides targeted guidance to refine the model’s output, with final results recorded accordingly. This setup allows us to assess the impact of human feedback on assertion quality and coverage. The third benchmark evaluates the model’s ability to autonomously improve coverage over multiple iterations without human intervention. By tracking coverage and assertion quality across five iterations, we aim to understand the effectiveness and limitations of the coverage improvement agent in a fully automated setting.

\begin{table}[!h]
    \caption{Key Performance Indicators of Saarthi on designs of various complexity}
    \renewcommand\arraystretch{1.2}
    \begin{center}
    \resizebox{\textwidth}{!}{%
    \begin{tabular}{cc*{3}{|ccc}}
        \toprule
        \multirow{2}*{\textbf{Design}} & \multirow{2}*{\textbf{Metric}} & \multicolumn{3}{c}{\textbf{Pass@1}} & \multicolumn{3}{c}{\textbf{Pass@2}} & \multicolumn{3}{c}{\textbf{Pass@3}} \\
        \cmidrule(lr){3-5} \cmidrule(lr){6-8} \cmidrule(lr){9-11}
        & & GPT-4.1 & GPT-5 & Llama3.3 & GPT-4.1 & GPT-5 & Llama3.3 & GPT-4.1 & GPT-5 & Llama3.3 \\
        \midrule
        \multirow{5}*{ECC} & \# Assertions & 19 & 28 & 14 & 25 & 32 & 6 & 25 & 48 & 12 \\
         & 1st generation & No & Yes & No & Yes & Yes & Yes & No & Yes & Yes \\
         & \# attempts to fix & 1 & 0 & 2 & 0 & 0 & 0 & 2 & 0 & 0 \\
         & \% Proven & 71.86\% & 89.28\% & 57.14\% & 64.00\% & 75.00\% & 33.33\% & 64.00\% & 81.25\% & 50.00\% \\
         & \% Coverage & 62.67\% & 58.57\% & 68.01\% & 60.21\% & 61.86\% & 2.18\% & 58.15\% & 93.74\% & 43.90\% \\
        \midrule
        \multirow{5}*{Automotive IP} & \# Assertions & 36 & 58 & 9 & 64 & 76 & 16 & 45 & 54 & 11 \\
         & 1st generation & No & No & No & No & No & No & No & No & No \\
         & \# attempts to fix & 4 & 3 & 5 & 5 & 3 & 5 & 3 & 3 & 5 \\
         & \% Proven & 50\% & 84.48\% & 22.22\% & 50\% & 56.57\% & 18.75\% & 48.88\% & 55.55\% & 45.45\% \\
         & \% Coverage & 64.29\% & 80.99\% & 6.99\% & 76.6\% & 77.02\% & 8.96\% & 72.75\% & 80.38\% & 42.44\% \\
        \midrule
        \multirow{5}*{Memory Scheduler~\cite{nvidiacvdp}} & \# Assertions & 24 & 22 & 13 & 16 & 28 & 7 & 23 & 29 & 17 \\
         & 1st generation & No & No & No & Yes & Yes & Yes & No & No & Yes \\
         & \# attempts to fix & 2 & 2 & 3 & 0 & 0 & 0 & 1 & 1 & 0 \\
         & \% Proven & 35.71\% & 40.91\% & 30.77\% & 50.00\% & 32.14\% & 71.43\% & 21.74\% & 32.01\% & 17.65\% \\
         & \% Coverage & 42.93\% & 54.73\% & 39.47\% & 43.32\% & 58.21\% & 44.20\% & 46.39\% & 48.86\% & 37.30\% \\
        \midrule
        \multirow{5}*{AXI4Lite~\cite{nvidiacvdp}} & \# Assertions & 69 & 139 & 39 & 75 & 87 & 60 & 74 & 117 & 92 \\
         & 1st generation & No & No & No & No & No & No & No & No & No \\
         & \# attempts to fix & 5 & 3 & 5 & 5 & 3 & 5 & 3 & 3 & 5 \\
         & \% Proven & 46.37\% & 72.66\% & 64.10\% & 40\% & 68.96\% & 45.00\% & 62.16\% & 61.15\% & 23.90\% \\
         & \% Coverage & 31.29\% & 44.06\% & 32.83\% & 35\% & 50\% & 36.42\% & 41.51\% & 41.18\% & 37.16\% \\
        \midrule
        \multirow{5}*{CIC Decimator~\cite{nvidiacvdp}} & \# Assertions & 16 & 36 & 10 & 16 & 37 & 10 & 19 & 30 & 10 \\
         & 1st generation & Yes & Yes & No & Yes & No & Yes & Yes & Yes & No \\
         & \# attempts to fix & 0 & 0 & 1 & 0 & 1 & 0 & 0 & 0 & 1 \\
         & \% Proven & 62.5\% & 75\% & 40\% & 31.25\% & 91.89\% & 30\% & 52.63\% & 86.67\% & 30\% \\
         & \% Coverage & 66.67\% & 74.45\% & 50\% & 52.14\% & 77.39\% & 18.22\% & 56.67\% & 72.86\% & 48.42\% \\
        \midrule
        \multirow{5}*{Float Multiplier} & \# Assertions & 19 & 49 & 10 & 17 & 42 & 0 & 23 & 57 & 15 \\
         & 1st generation & Yes & Yes & Yes & Yes & Yes & No & Yes & No & Yes \\
         & \# attempts to fix & 0 & 0 & 0 & 0 & 0 & 3 & 0 & 2 & 0 \\
         & \% Proven & 10.53\% & 51.02\% & 20\% & 23.53\% & 30.95\% & 0\% & 17.39\% & 42.11\% & 6.67\% \\
         & \% Coverage & 4.97\% & 65.37\% & 12.14\% & 19.96\% & 83.71\% & 0\% & 9.44\% & 78.48\% & 3.15\% \\
        \bottomrule
        \\[-1.5ex]
    \end{tabular}
    }
    \scriptsize
    \vspace{2ex} 
    \label{basic_results}
    \end{center}
\end{table}

%\begin{figure}[h!]
%\centering
%  \begin{subfigure}{0.32\textwidth}
%    \centering
%    \includegraphics[width=\textwidth]{Images/radar_chart_pass_1.pdf}
%    \caption{Pass@1 for different designs}
%    \label{fig:radar_chart_pass_1}
%  \end{subfigure}
%  \begin{subfigure}{0.32\textwidth}
%    \centering
%    \includegraphics[width=\textwidth]{Images/radar_chart_pass_2.pdf}
%    \caption{Pass@2 for different designs}
%    \label{fig:radar_chart_pass_2}
%  \end{subfigure}
%  \begin{subfigure}{0.32\textwidth}
%    \centering
%    \includegraphics[width=\textwidth]{Images/radar_chart_pass_3.pdf}
%    \caption{Pass@3 for different designs}
%    \label{fig:radar_chart_pass_3}
%  \end{subfigure}
%\caption{Radar charts for Pass@1, Pass@2, and Pass@3 for different designs}
%\label{fig:radar_charts}
%\end{figure}

\begin{figure}[h!]
\centering
  \includegraphics [width=0.5\textwidth] {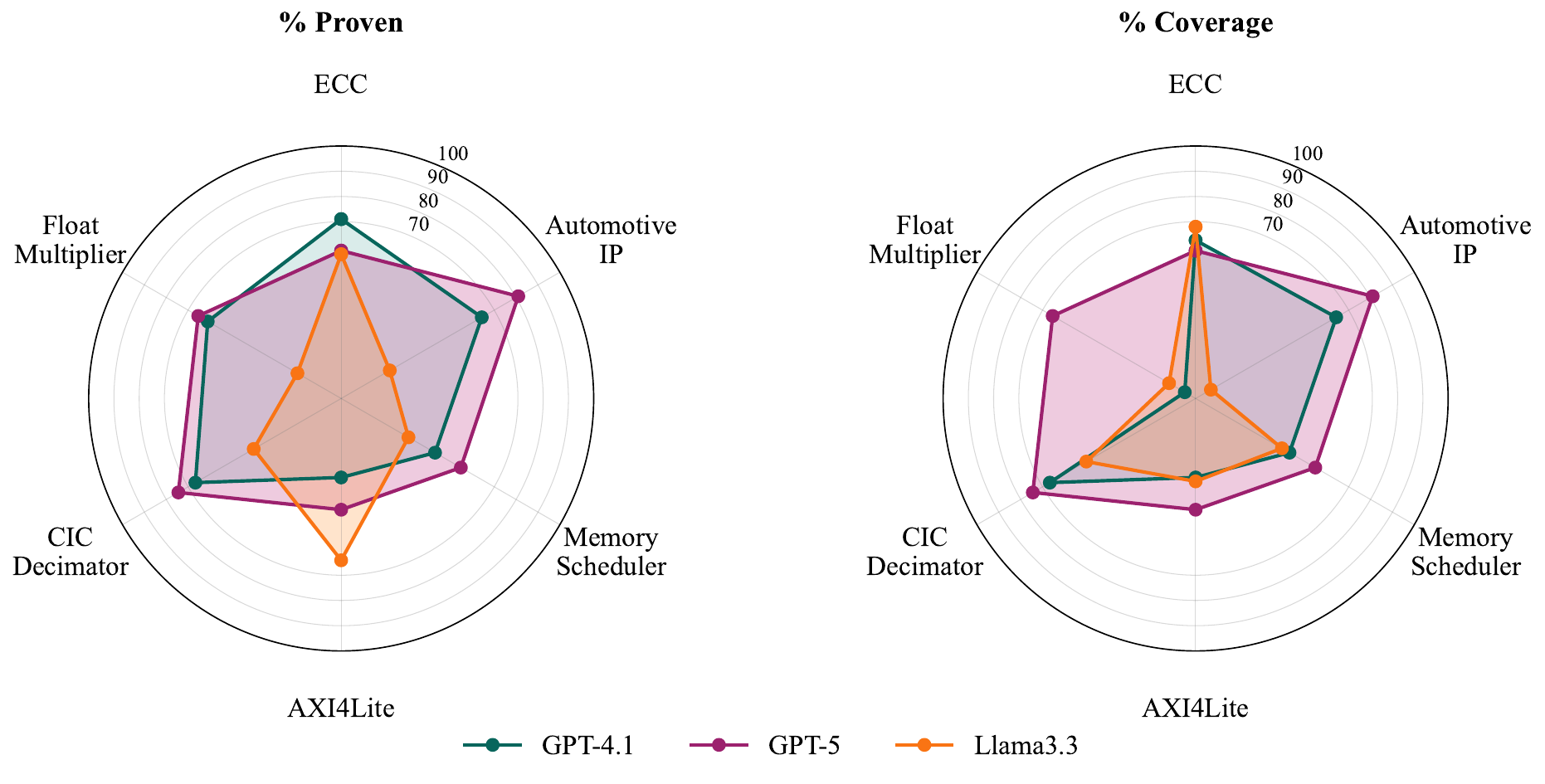}
\caption{Radar chart for Pass@1 for different designs}
\label{fig:radar_chart_pass_1}
\end{figure}

\begin{figure}[h!]
\centering
  \includegraphics [width=0.5\textwidth] {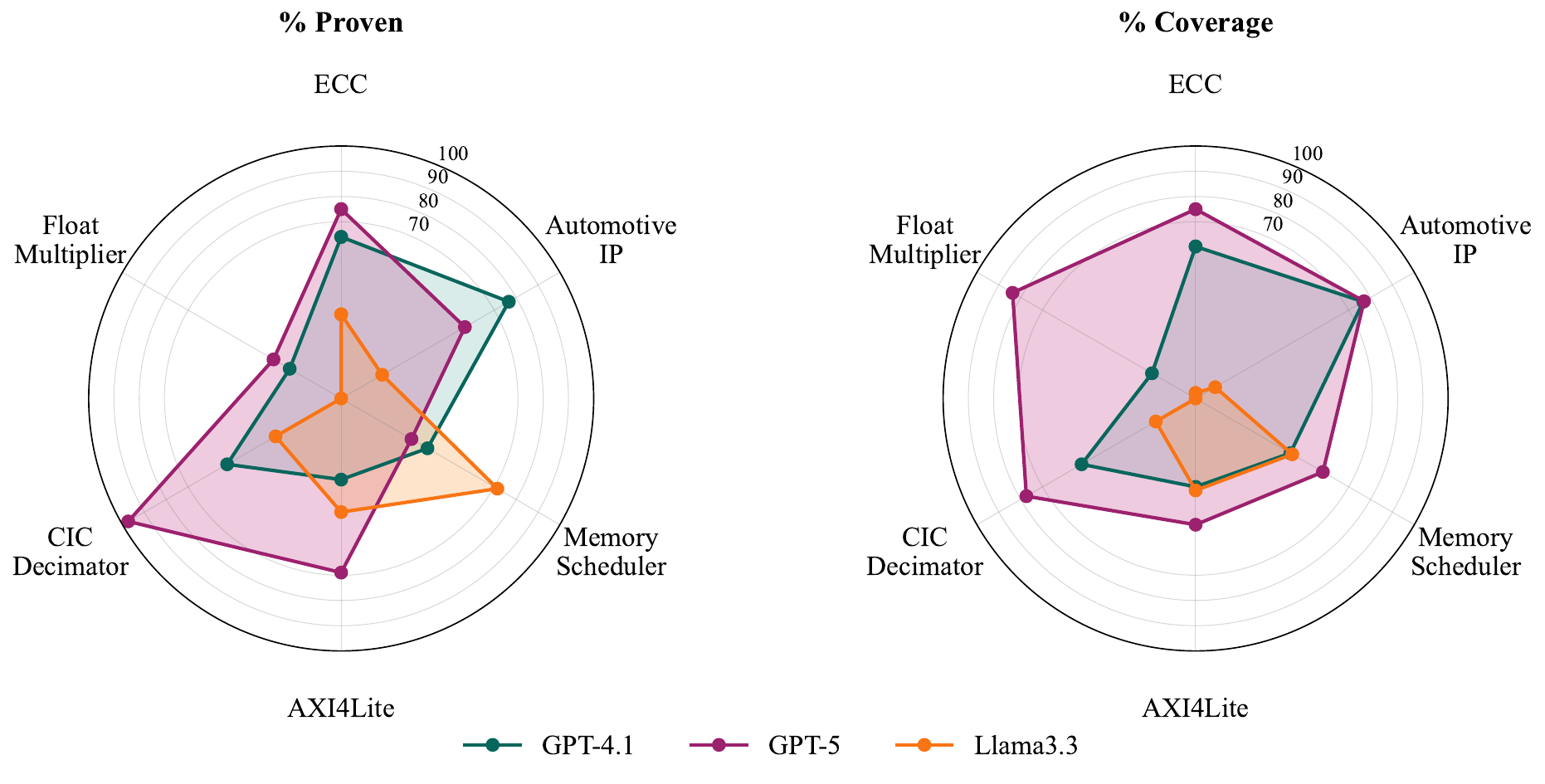}
\caption{Radar chart for Pass@2 for different designs}
\label{fig:radar_chart_pass_2}
\end{figure}

\begin{figure}[h!]
\centering
  \includegraphics [width=0.5\textwidth] {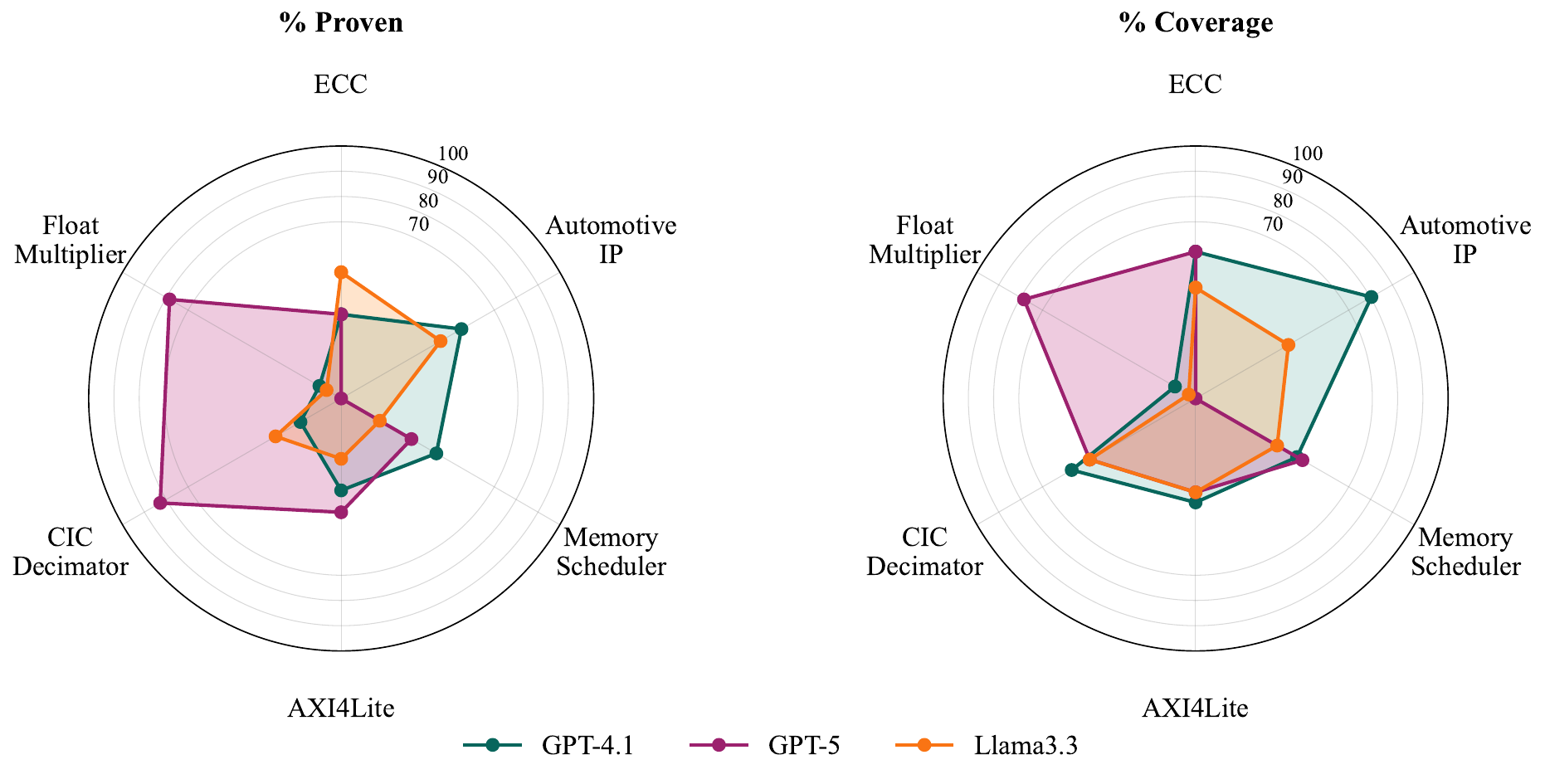}
\caption{Radar chart for Pass@3 for different designs}
\label{fig:radar_chart_pass_3}
\end{figure}

The results in Table~\ref{basic_results} present the \acp{KPI} of Saarthi. The results show that Saarthi can generate formal assertions for a wide range of hardware designs. GPT-4.1 performs well on simpler modules, while GPT-5 significantly improves both assertion proof rates and formal coverage for complex designs like the Float Multiplier. This improvement comes at the cost of higher latency, as GPT-5 uses a reasoning-based approach that requires more processing time. This trade-off between accuracy and speed is critical for practical deployment. Additionally, specification quality strongly impacts results: simple specs work for basic designs (e.g., ECC, CIC decimator), but vague natural language descriptions in complex RTL lead to divergence from ideal assertions. For example, the entire CVDP Memory Scheduler spec contained only a few lines of expected states. In real-world scenarios, continuous collaboration between design and verification teams is essential, so despite minimal guidance and limited specification, we believe the LLMs achieved impressive outcomes.

\begin{table}[!h]
    \caption{Coverage improvement with and without \ac{HIL} at Pass@1}
    \renewcommand\arraystretch{1.2}
    \begin{center}
    \resizebox{\textwidth}{!}{%
    \begin{tabular}{cc|ccc|ccc}
        \toprule
        \multirow{2}*{\textbf{Design}} & \multirow{2}*{\textbf{Metric}} & \multicolumn{3}{c|}{\textbf{Without HIL}} & \multicolumn{3}{c}{\textbf{With HIL}} \\
        \cmidrule(lr){3-5} \cmidrule(lr){6-8}
        & & GPT-4.1 & GPT-5 & Llama3.3 & GPT-4.1 & GPT-5 & Llama3.3 \\
        \midrule
        \multirow{3}*{ECC} & \# Assertions & 53 & 66 & 19 & 53 & 60 & 19 \\
         & \% Proven & 84.90\% & 78.80\% & 68.42\% & 96.06\% & 80.00\% & 84.21\% \\
         & \% Coverage & 89.31\% & 92.83\% & 90.63\% & 98.05\% & 95.22\% & 92.53\% \\
        \midrule
        \multirow{3}*{Automotive IP} & \# Assertions & 36 & 58 & 19 & 36 & 58 & 19 \\
         & \% Proven & 50\% & 84.48\% & 26.31\% & 72.22\% & 93.10\% & 63.15\% \\
         & \% Coverage & 64.29\% & 80.99\% & 38.87\% & 73.91\% & 85.94\% & 64.91\% \\
        \midrule
        \multirow{3}*{Memory Scheduler} & \# Assertions & 27 & 35 & 36 & 27 & 35 & 36 \\
         & \% Proven & 62.96\% & 48.57\% & 33.33\% & 85.19\% & 57.14\% & 55.56\% \\
         & \% Coverage & 42.35\% & 50.00\% & 41.11\% & 61.54\% & 62.02\% & 45.65\% \\
        \midrule
        \multirow{3}*{AXI4Lite} & \# Assertions & 101 & 139 & 3 & 101 & 139 & 3 \\
         & \% Proven & 70.29\% & 72.66\% & 33.33\% & 85.10\% & 83.45\% & 66.66\% \\
         & \% Coverage & 38.90\% & 44.06\% & 28.23\% & 61.56\% & 67.06\% & 31.12\% \\
        \midrule
        \multirow{3}*{CIC Decimator} & \# Assertions & 28 & 53 & 28 & 28 & 53 & 28 \\
         & \% Proven & 67.86\% & 81.13\% & 50\% & 100\% & 100\% & 100\% \\
         & \% Coverage & 79.69\% & 73.14\% & 55.64\% & 91.20\% & 91.60\% & 91.32\% \\
        \midrule
        \multirow{3}*{Float Multiplier} & \# Assertions & 88 & 126 & 10 & 88 & 126 & 10 \\
         & \% Proven & 79.55\% & 61.91\% & 20\% & 100\% & 100\% & 100\% \\
         & \% Coverage & 27.46\% & 64.72\% & 12.23\% & 67.57\% & 92.83\% & 74.68\% \\
        \bottomrule
        \\[-1.5ex]
    \end{tabular}
    }
    \scriptsize
    \vspace{2ex} 
    \label{hil_coverage_results}
    \end{center}
\end{table}

Table~\ref{hil_coverage_results} illustrates the impact of \ac{HIL} collaboration in reducing the challenges of interpreting natural language specifications. By improving specification visibility and providing timely feedback, the models achieved higher assertion quality and coverage compared to the initial results. This interaction not only improved accuracy but also significantly reduced the time required, transforming what would typically take hours of manual effort into a much faster, collaborative process. These findings indicate that, even in their current state, LLMs can substantially enhance productivity by serving as intelligent assistants, allowing engineers to focus on complex reasoning tasks while delegating repetitive steps to AI.

\begin{table}[!h]
    \caption{Coverage improvement without HIL across iterations (GPT-5)}
    \renewcommand\arraystretch{1.2}
    \begin{center}
    \resizebox{0.65\textwidth}{!}{% Scale table to 90% of text width
    \begin{tabular}{cc*{5}{|c}}
        \toprule
        \multirow{2}*{\textbf{Design}} & \multirow{2}*{\textbf{Metric}} & \multicolumn{5}{c}{\textbf{Coverage improvement without HIL}} \\
        \cmidrule(lr){3-7}
        & & 1\textsuperscript{st} Iter & 2\textsuperscript{nd} Iter & 3\textsuperscript{rd} Iter & 4\textsuperscript{th} Iter & 5\textsuperscript{th} Iter \\
        \midrule
        \multirow{3}*{ECC} & \# Assertions & 312 & 318 & 1258 & N/A & N/A \\
         & \% Proven & 91.67\% & 92.12\% & 97.93\% & N/A & N/A \\
         & \% Coverage & 88.89\% & 86.86\% & 87.87\% & N/A & N/A \\
        \midrule
        \multirow{3}*{Automotive IP} & \# Assertions & 59 & 71 & 134 & 185 & 233 \\
         & \% Proven & 81.35\% & 77.46\% & 85.82\% & 87.75\% & 89.69\% \\
         & \% Coverage & 80.56\% & 79.95\% & 75.14\% & 74.31\% & 72.69\% \\
        \midrule
        \multirow{3}*{Memory Scheduler} & \# Assertions & 35 & 36 & 37 & 37 & 37 \\
         & \% Proven & 40.00\% & 41.12\% & 43.24\% & 42.50\% & 43.80\% \\
         & \% Coverage & 55.64\% & 56.10\% & 55.85\% & 56.40\% & 55.20\% \\
\midrule
        \multirow{3}*{AXI4Lite} & \# Assertions & 104 & 145 & 165 & 199 & 232 \\
         & \% Proven & 62.50\% & 58.62\% & 59.39\% & 59.29\% & 59.99\% \\
         & \% Coverage & 45.36\% & 45.36\% & 47.21\% & 52.99\% & 48.93\% \\
        \midrule
        \multirow{3}*{CIC Decimator} & \# Assertions & 53 & 63 & 86 & 116 & 203 \\
         & \% Proven & 81.13\% & 79.37\% & 80.23\% & 81.90\% & 80.30\% \\
         & \% Coverage & 73.14\% & 74.77\% & 72.34\% & 72.44\% & 64.37\% \\
        \midrule
        \multirow{3}*{Float Multiplier} & \# Assertions & 126 & 148 & 201 & 244 & 264 \\
         & \% Proven & 61.91\% & 57.43\% & 67.66\% & 60.25\% & 58.71\% \\
         & \% Coverage & 64.72\% & 71.60\% & 75.09\% & 76.84\% & 82.45\% \\
        \bottomrule
        \\[-1.5ex]
    \end{tabular}
    }
    \scriptsize
    \vspace{2ex} 
    %\parbox{0.65\textwidth}{ 
    %    \textit{Note: The data in this table is still in the process of being collected and finalized.}
    %}
    \label{hil_iterations_results}
    \end{center}
\end{table}

Table~\ref{hil_iterations_results} presents the progression of assertion quality and coverage across multiple iterations of the coverage improvement agent operating in a fully autonomous setting. Some designs, such as Automotive IP and Float Multiplier, show steady improvements in assertion count and coverage, whereas others, like Memory Scheduler and AXI4Lite, reach early saturation, with minimal gains beyond the initial iterations. These trends indicate that repeated autonomous runs do not always provide significant benefits, particularly when early iterations already achieve stable results. Interestingly, GPT-5 was able to generate thousands of assertions in a single run for ECC before crashing due to the large token requirements. This result highlights the practical limits of GPT-5 when handling extremely large assertion sets, even for moderately complex designs. In practice, a more efficient approach is to perform one autonomous run and then apply \ac{HIL} refinement to close remaining gaps.

\begin{figure}[h!]
\centering
  \includegraphics [width=0.7\textwidth] {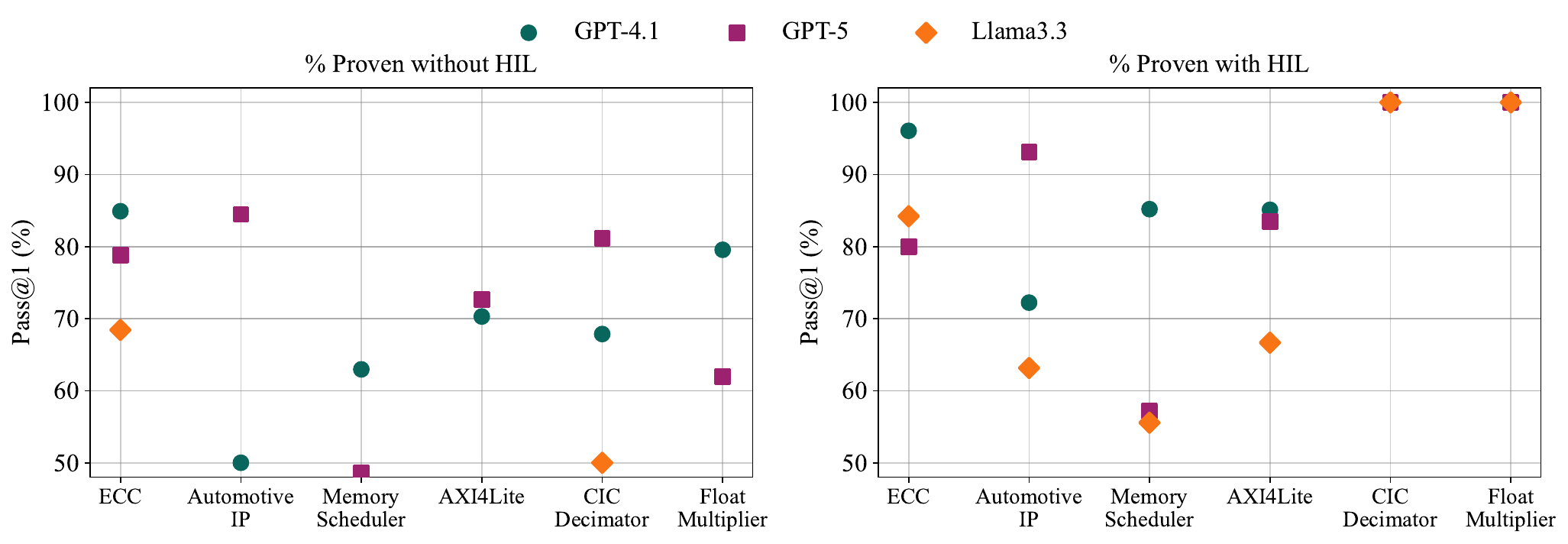}
\caption{\% Proven comparison with and without \ac{HIL} at Pass@1}
\label{fig:proven_comparison_pass1}
\end{figure}

\begin{figure}[h!]
\centering
  \includegraphics [width=0.7\textwidth] {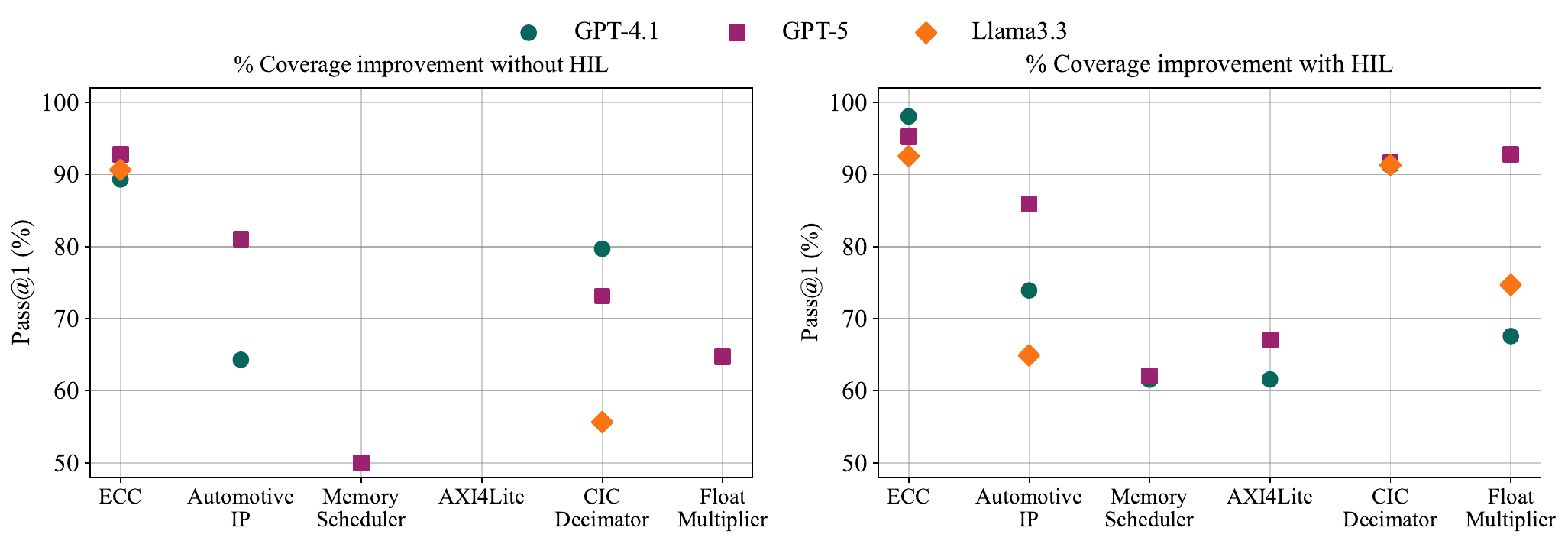}
\caption{Coverage comparison with and without \ac{HIL} at Pass@1}
\label{fig:coverage_comparison_pass1}
\end{figure}

\section{Conclusion} \label{conc}

The increasing complexity and safety-critical demands of semiconductor design have intensified the need for scalable, reliable AI-driven formal verification. While multi-agent collaboration and orchestration offer promising directions, challenges in controllability and adaptive learning remain. Saarthi addresses these limitations by combining structured assertion generation, domain-specific knowledge retrieval, and iterative coverage refinement to move toward \ac{DSGI} in formal verification. Across six \ac{RTL} designs, Saarthi demonstrated consistent improvements in assertion quality, proof convergence, and coverage. In particular, on NVIDIA CVDP designs such as AXI4Lite and CIC Decimator, the system achieved over 50\% and 77\% coverage, respectively, for first generation, with further gains observed through human-in-the-loop refinement. GPT-5 delivered the strongest results, especially on complex designs, though its advanced reasoning introduced moderate latency compared to smaller models. These findings validate the effectiveness of our rulebook-guided generation, hybrid retrieval strategies, and feedback-driven repair mechanisms. Similar to semiconductors, Moore's Law-like trends apply to \ac{AI}: over time, access becomes faster, cheaper, and more widespread. As benchmarking continues, Saarthi offers a concrete and practical step toward more dependable and efficient \ac{AI} integration towards \ac{DSGI}.

\section*{Acknowledgement}

% \hspace{11cm}With funding from the:

\begin{figure}[h!]
\centering
\begin{minipage}{.5\textwidth}
  \centering
  \includegraphics[width=.7\linewidth]{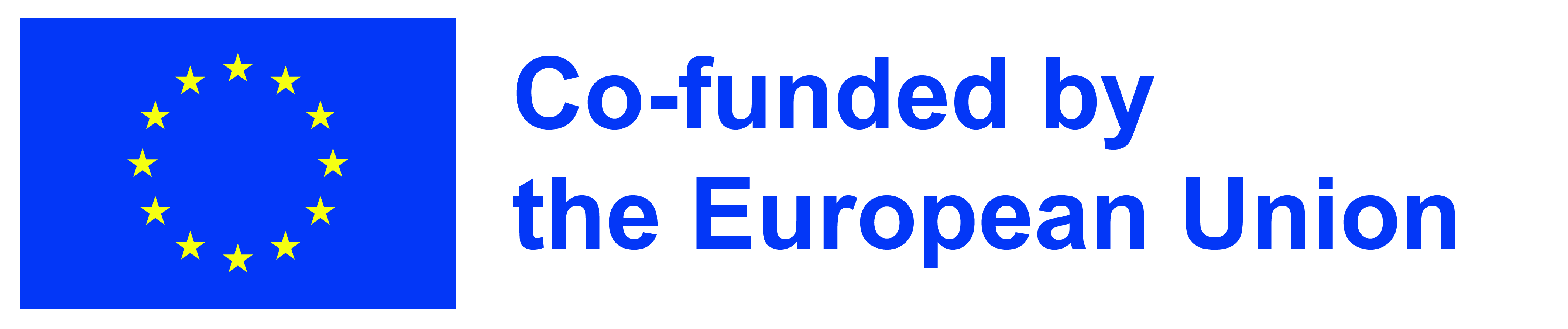}
\end{minipage}%
\begin{minipage}{.5\textwidth}
  \centering
  \includegraphics[width=.5\linewidth]{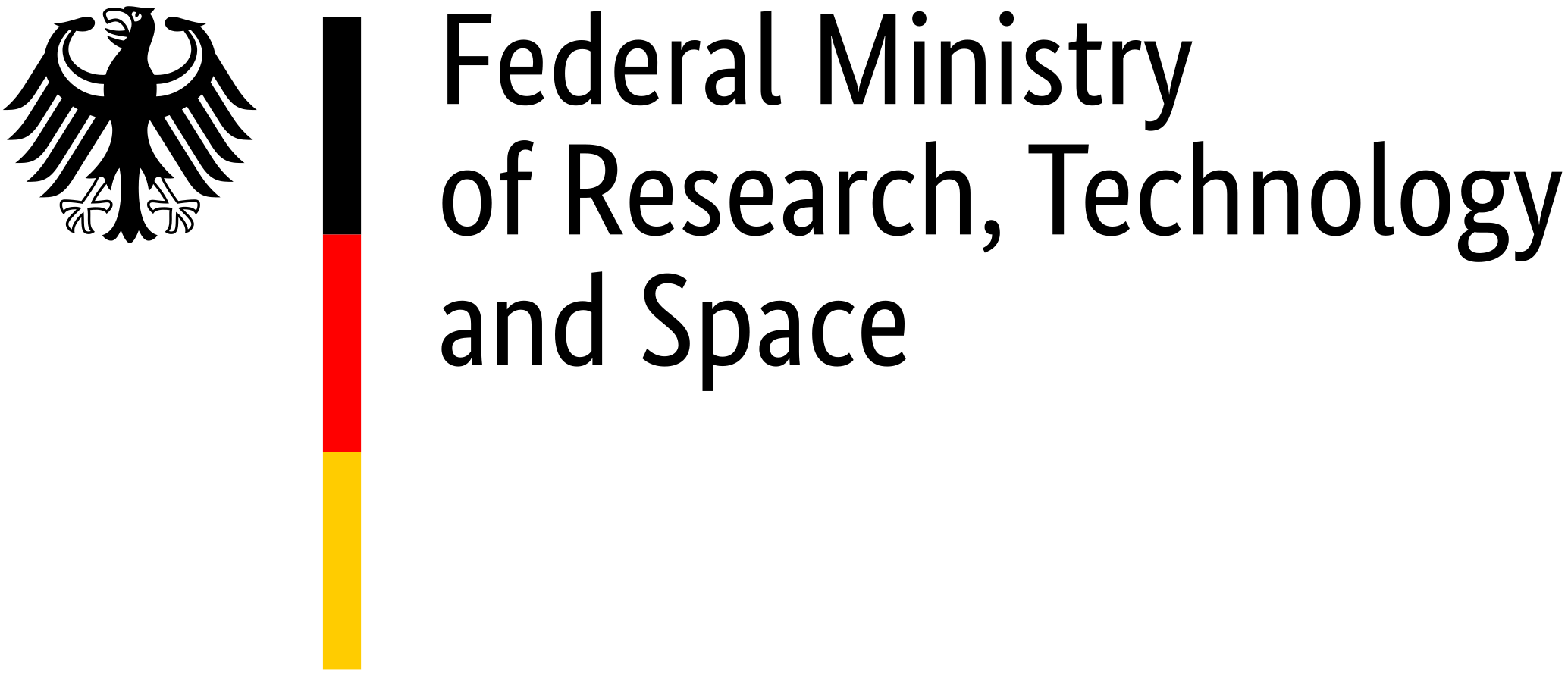}
\end{minipage}
\end{figure}

Under grant 101194371, Rigoletto is supported by the Chips Joint Undertaking and its members, including the top-up funding by the National Funding Authorities from involved countries.

Rigoletto is also funded by the Federal Ministry of Research, Technology and Space under the funding code 16MEE0548S. The responsibility for the content of this publication lies with the author.

\printbibliography[heading=bibintoc]

\end{document}